\documentclass[10pt,twocolumn,letterpaper]{article}

\usepackage[pagenumbers]{cvpr} 
\usepackage{algorithm}
\usepackage{algorithmic}
\usepackage{amsmath}
\usepackage{amsfonts}
\usepackage{amssymb}
\usepackage{booktabs}
\usepackage{multirow}
\usepackage{xcolor}
\usepackage[table]{xcolor}
\usepackage{colortbl}

\usepackage[american]{babel}
\usepackage{microtype}

\definecolor{cvprblue}{rgb}{0.21,0.49,0.74}
\usepackage[pagebackref,breaklinks,colorlinks,allcolors=cvprblue]{hyperref}

\title{Text-guided Feature Disentanglement for Cross-modal Gait Recognition}

\author{
Zhiyang Lu \quad Ming Cheng\thanks{Corresponding author.} \\
Fujian Key Laboratory of Urban Intelligent Sensing and Computing, Xiamen University. \\
Key Laboratory of Multimedia Trusted Perception and Efficient Computing, \\
Ministry of Education of China, Xiamen University, 361005, P.R. China.\\
}

\begin{document}
\maketitle
\begin{abstract}
Gait recognition is a biometric technique that identifies individuals based on their walking patterns, offering advantages in long-range, non-intrusive scenarios. However, real-world scenarios often involve heterogeneous sensing modalities such as LiDAR and RGB cameras, making LiDAR-Camera Cross-modal Gait recognition (LCCGR) a critical yet challenging task due to the substantial modality gap between 2D videos and 3D point cloud sequences. To address this challenge, we propose TCFDNet, a Text-guided Cross-modal Feature Disentanglement Network, which leverages modality-aware textual priors as semantic anchors to guide the learning of disentangled modality-shared representations. Specifically, we construct a Gait Modality Text Dictionary (GMTD) using large language models to generate rich semantic descriptions of gait across modalities and viewpoints. A CLIP-based Multi-grained Feature Encoder then aligns visual and textual features within a unified vision-language space. Furthermore, the Text-guided Feature Disentanglement (TFD) module selects the $top\text{-}k$ matched textual descriptions to reconstruct modality-specific representations and derive modality-shared features via residual decomposition and orthogonality constraints. To mitigate the fragility of the disentangled shared features, we propose a Feature Stability Enhancement (FSE) module, which models spatial and channel-wise correlations to improve feature robustness. In addition, a cross-modal patch exchange strategy is introduced to further improve generalization. Extensive experiments on SUSTech1K and FreeGait datasets demonstrate that TCFDNet achieves new state-of-the-art results and validate the effectiveness of the proposed modules.
\end{abstract}
\section{Introduction}
\label{sec:intro}
\begin{figure}[t]
\centering
\includegraphics[width=\linewidth]{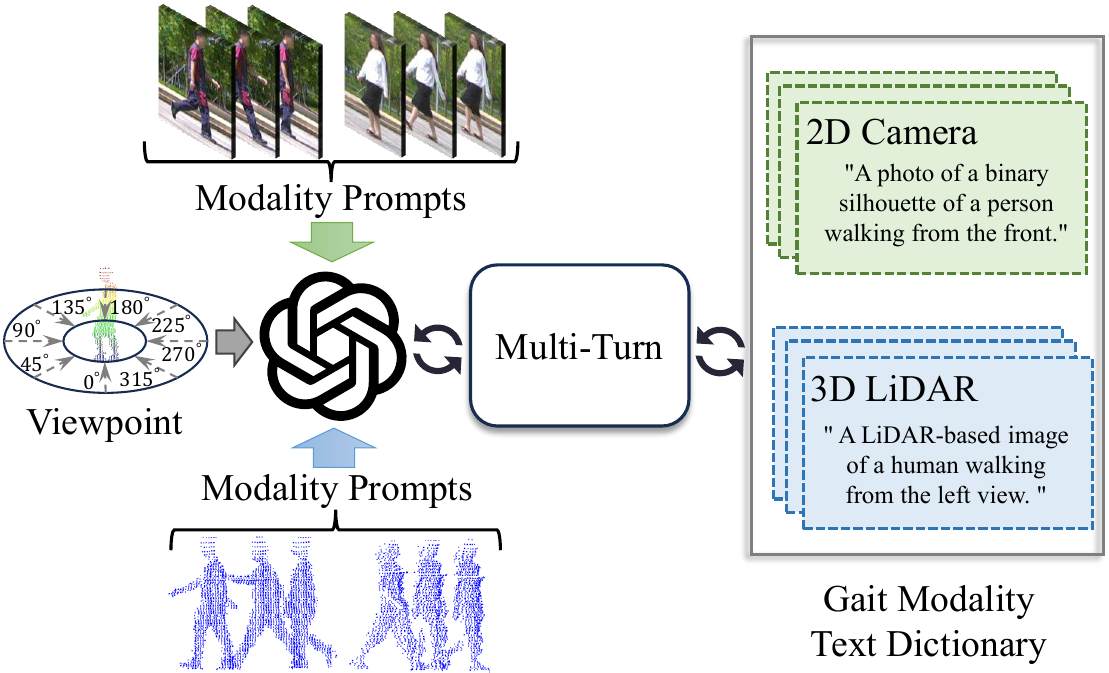}
\caption{Details of the GMTD construction.}
\label{fig::gmtd}
\vspace{-0.6cm}
\end{figure}
Gait recognition aims to identify individuals based on their walking patterns. Compared with facial and iris recognition, gait offers significant advantages such as being non-contact, long-range, and being difficult to disguise, which endow it with broad application prospects in intelligent surveillance, suspect tracking, and health diagnostics \cite{jin2025exploring-multigait-aaai,shen2024comprehensive-survey-gait-deeplearning,fan2023opengait-deeplearning-opengait,chao2019gaitset-AAAI-appearance-deeplearning,shen2024comprehensive-gait-survey}. Although 2D and 3D single-modality gait recognition methods have achieved remarkable performance~\cite{shen2024comprehensive-gait-survey,dong2022survey-in-context-learning,yang2025bridging-gait-cvpr,jin2025denoisinggait-cvpr,shen2025lidargait++-cvpr,shen2023lidargait-3d-lidargait,han2024gait-3d-freegait}, the proliferation of diverse sensors has increasingly highlighted the need for cross-modality collaboration~\cite{wang2025tokenmatcher-vireid-aaai,yu2023modality-VIreid-cross-modal-balance,huang2025l4dr-aaai-radarlidar,huang2025v2x-radarlidar}. 
Consequently, LiDAR-Camera Cross-modality Gait Recognition (LCCGR) has emerged as a critical component for multi-device collaborative retrieval. \textit{Owing to the significant heterogeneity between 2D camera videos and 3D LiDAR point clouds sequences, LCCGR faces two principal challenges: (1) bridging the modality gap between 2D and 3D inputs, and (2) extracting discriminative gait features in the modality-shared space.} Guo et al. constructed a synthetic 2D-3D dataset using RGB images and employed contrastive learning loss for model pretraining, followed by fine-tuning on specific datasets \cite{guo2025camera-crossmodal-CLGait}. Although this approach achieved promising results, the domain gap between synthetic and real-world data introduced biases into the model. Wang et al. attempted to learn modality-shared prototype features and applied attention mechanisms for adaptive weighting \cite{wang2024cross-crossmodal-CrossGait}. However, the prototype lacked effective generalization capability. Furthermore, existing feature disentanglement networks are essentially unreliable and interpretable, acting as a black box~\cite{zhu2021and-Intro-CVPR2021-interpretable,bass2020icam-Intro-NeurIPS-interpretable}. The advent of vision-language models, such as CLIP~\cite{radford2021learning-clip}, offers a novel perspective for extracting modality-shared features: by describing modality-specific information through text and projecting it into the visual space, so as to guide feature disentanglement reliably.

Motivated by this insight, we propose a \textbf{Text-guided Cross-modal Feature Disentanglement Network (TCFDNet)}, which leverages modality-specific textual priors to decouple modality-specific and modality-shared gait representation. We pre-construct the Gait Modality Text Dictionary (GMTD) to encode modality-specific textual priors as semantic anchors for cross-modal alignment. Specifically, by leveraging large language models (LLMs), GMTD generates modality-aware text descriptions enriched with multi-view information from eight predefined viewpoints. To enhance semantic diversity, we adopt an $l$-round multi-turn interaction strategy with the LLMs (Fig.~\ref{fig::gmtd}) to scale it up to a larger magnitude. To this end, upon obtaining the textual modality descriptions, we project both the textual and visual information into a shared vision-language space. Specifically, we design a multi-grained feature encoder built upon CLIP. The textual descriptions and gait visual data are first processed by the CLIP text encoder and the visual encoder, respectively, to obtain modality-specific textual embeddings and global gait representations within a vision-language shared space. Furthermore, to enhance fine-grained feature extraction, we freeze the CLIP weights and introduce a lightweight ResNet-based bypass adapter that captures fine-grained gait features. Finally, a Multi-grained Fusion (MF) module integrates these visual features across multiple granularities and aligns them with the textual representations.

To achieve text-driven cross-modal representation disentanglement, we introduce a Text-guided Feature Disentanglement (TFD) module that explicitly decouples modality-shared and modality-specific information. Concretely, we first leverage the image–text similarity matrix produced by the CLIP encoder to retrieve the $top\text{-}k_{t}$ textual descriptions from the GMTD that are most semantically aligned with the current visual representation. The semantic embeddings of these highly relevant texts are subsequently utilized to guide the reconstruction of the modality-specific component.
Furthermore, a residual decomposition strategy combined with an orthogonality constraint is employed to enforce statistical independence between the shared and specific representations, thereby yielding a well-structured and explicitly disentangled feature space.

Nevertheless, the modality-shared representations used for retrieval—derived solely from residual decomposition—tend to suffer from limited robustness, as they are highly sensitive to feature perturbations~\cite{chen2024rodla-Intro-CVPR2024-robustness}. To alleviate this limitation, we introduce a Feature Stability Enhancement (FSE) module that explicitly regulates shared representations from two complementary perspectives: spatial dependency and channel-wise correlation. We further introduce a cross-modal Patch Exchange augmentation strategy that swaps localized regions between heterogeneous modalities while preserving semantic consistency. This simple yet effective technique significantly enhances the adaptability to cross-modal discrepancies during training. In summary, our contributions are as follows:

\begin{itemize}
    \item We propose a cross-modal gait recognition framework that leverages modality-specific text descriptions to disentangle visual representations, thereby capturing shared cues. To achieve this, we design a CLIP-based multi-grained encoder and a Multi-grained Fusion (MF) module to integrate both local and global information effectively.

    \item A Text-guided Feature Disentanglement (TFD) module is developed to reconstruct modality-specific semantics from text and disentangle modality-shared representations via residual decomposition and orthogonality constraints.

    \item A Feature Stability Enhancement (FSE) module is proposed to improve the robustness of modality-shared features by modeling spatial dependencies and inter-channel relationships.

    \item Our method achieves state-of-the-art performance on multiple cross-modal gait benchmarks, with extensive ablations confirming the contribution of each component.
\end{itemize}
\begin{figure*}[t]
\centering
\includegraphics[width=\textwidth]{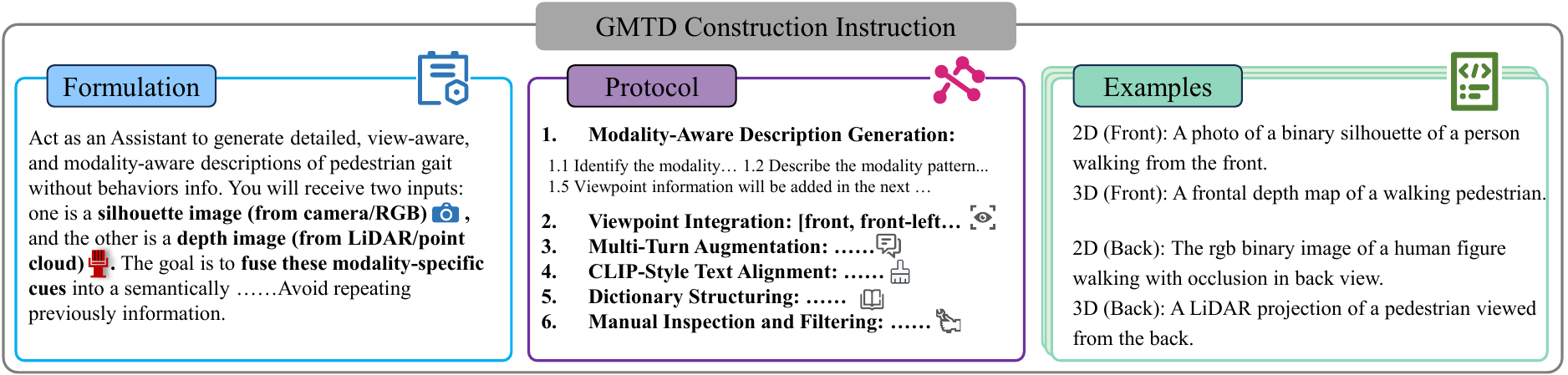}
\caption{The instruction for GMTD, which consists of three parts: formulation, protocol, and examples. This design encourages LLMs\cite{achiam2023gpt,bai2023qwen-report,bai2023qwen-vl} to perform instruction-following~\cite{lou2023comprehensive-instruct-following}, chain-of-thought\cite{wei2022chain-chain-of-thought}, and in-context generation~\cite{dong2022survey-in-context-learning}.}
\label{fig::prompt}
\vspace{-0.3cm}
\end{figure*}

\section{Related Work}
\subsection{Gait Recognition}
Gait recognition has long been studied under two major paradigms: 2D camera-based and 3D LiDAR-based approaches. 2D methods are commonly divided into appearance-based and model-based branches. Appearance-based methods directly exploit silhouettes or RGB sequences to capture motion patterns without relying on explicit body models. Early works introduced handcrafted descriptors such as Gait Energy Images~\cite{han-2005-tpami-gaitenergy} or temporal templates~\cite{wang2010chrono-template}, while recent deep learning approaches extract fine-grained spatiotemporal features using CNN- or Transformer-based backbones~\cite{chao2019gaitset-AAAI-appearance-deeplearning,fan2020gaitpart-appearance-deeplearning,lin2021gait-appearance-deeplearning-ICCV-celoss,jin2025exploring-multigait-aaai,jin2025denoisinggait-cvpr,yang2025bridging-gait-cvpr}. These methods often aggregate per-frame motion cues to construct global gait embeddings. Model-based methods rely on human pose estimation to represent gait as a sequence of joint locations~\cite{wang2023gait-TMM-skeleton,li2022strong-cyclegait-TMM-Skeleton,fan2024skeletongait-gait-skeleton}. Graph convolutional networks~\cite{teepe2021gaitgraph-skeleton-graph} and temporal attention mechanisms have been proposed to better model structured motion dynamics. However, camera-based methods are vulnerable to variations in viewpoint, clothing, and illumination. To overcome these limitations, LiDAR-based gait recognition has recently emerged as a robust alternative. Unlike RGB cameras, LiDAR sensors provide geometry-rich, illumination-invariant data. LidarGait~\cite{shen2023lidargait-3d-lidargait} demonstrated the superiority of LiDAR over camera in challenging scenarios. FreeGait~\cite{han2024gait-3d-freegait} further extended this line of research by collecting large-scale gait data in outdoor, in-the-wild environments, leveraging both point cloud and range-view representations.

\subsection{Cross-Modal Gait Recognition}
While unimodal and multimodal gait recognition have achieved remarkable progress~\cite{han-2005-tpami-gaitenergy,wang2010chrono-template,chao2019gaitset-AAAI-appearance-deeplearning,fan2020gaitpart-appearance-deeplearning,lin2021gait-appearance-deeplearning-ICCV-celoss,han2024gait-3d-freegait,fan2024skeletongait-gait-skeleton,fu2023gpgait-iccv,zhu2023gaitref-ijcb,zou2024multi-multimodal-adptive-gait,jin2025exploring-multigait-aaai,jin2025denoisinggait-cvpr,yang2025bridging-gait-cvpr,teepe2021gaitgraph-skeleton-graph,fan2025opengait-tpami}, cross-modal gait recognition—especially between LiDAR and camera modalities—remains underexplored. The primary challenge lies in the large modality gap, often exceeding intra-class variation, making feature alignment across modalities difficult. Several efforts have attempted to bridge this gap. CL-Gait~\cite{guo2025camera-crossmodal-CLGait} pre-trains encoders with contrastive learning on synthetic 2D/3D data to unify modality-specific spaces. However, this approach heavily relies on synthetic datasets and struggles to generalize to real-world distributions. CrossGait~\cite{wang2024cross-crossmodal-CrossGait} proposes learnable shared prototypes to align heterogeneous features. Yet, directly pulling features from different modalities together often leads to class collapse and reduced discriminability. Inspired by the success of vision-language models such as CLIP~\cite{radford2021learning-clip}, recent works have explored the potential of textual semantics in bridging heterogeneous modalities. However, existing gait recognition frameworks rarely exploit language as a supervisory or regularizing signal. In this work, we propose a novel framework that leverages LLMs to generate modality-specific textual descriptions, which are embedded into a shared visual-language space via CLIP. These textual features are then used to decouple modality-specific visual information, serving as anchors to guide the disentanglement and alignment of features.
\section{Methodology}
\subsection{Overview}
\begin{figure*}[t]
\centering
\includegraphics[width=\textwidth]{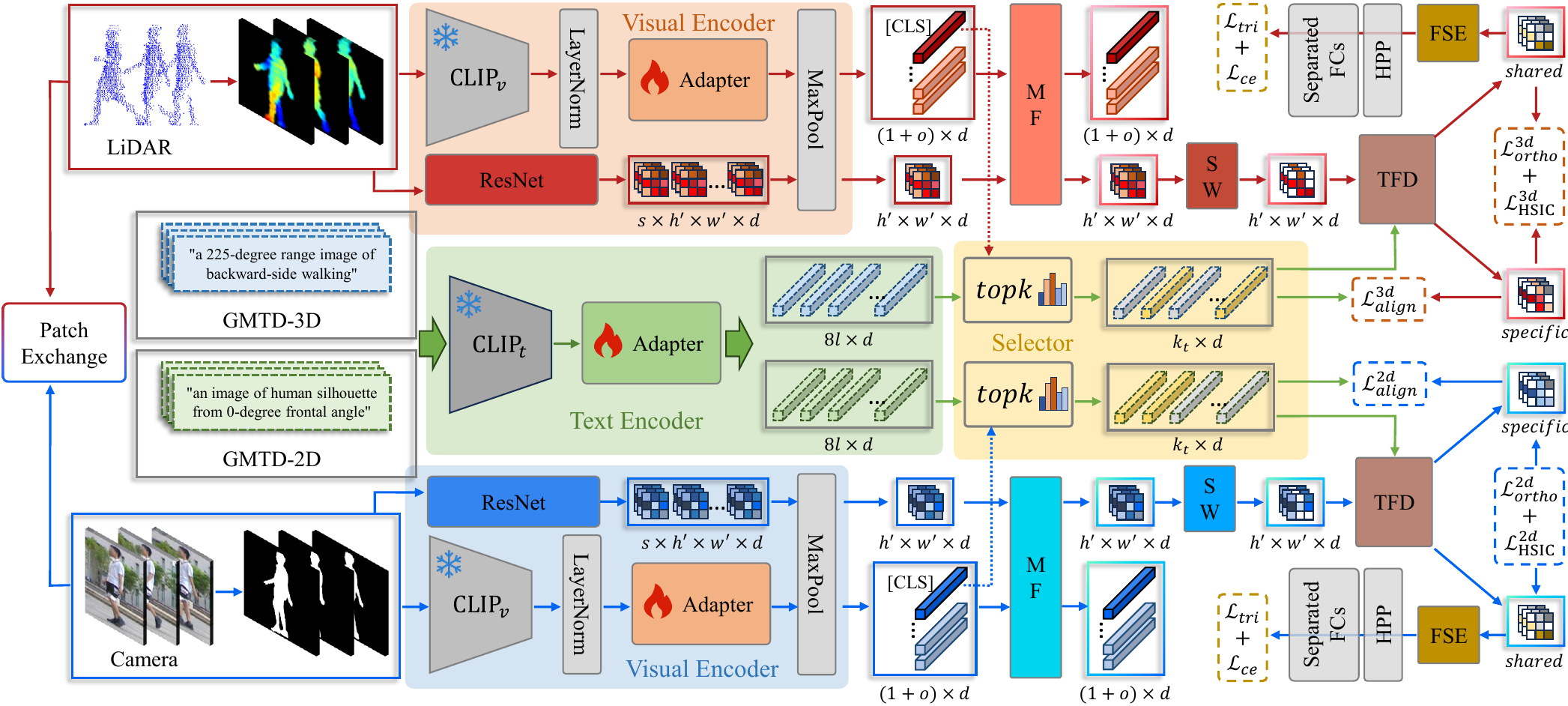}
\caption{Illustration of the proposed framework.}
\label{fig::framework}
\vspace{-0.3cm}
\end{figure*}

Inspired by the cross-modal alignment capabilities of vision-language models between text and visual spaces, we propose TCFDNet, which leverages modality-specific descriptions of the gait dictionary to disentangle modality-shared features for the LCCGait task. The overall framework is illustrated in Figure~\ref{fig::framework}. We propose a patch exchange data augmentation strategy that switches localized areas between modalities at the input level to improve the cross-modal perception, which is analyzed in the Supplementary Material. Meanwhile, we construct a diverse dictionary of gait modality-specific texts using large language models (LLMs), and then design a CLIP-based multi-grained feature encoder to extract both modality-specific textual embeddings and multi-grained gait visual features. Subsequently, the Text-guided Feature Disentanglement (TFD) module leverages the aligned modality-specific textual priors to reconstruct and disentangle the original visual features, producing shared cross-modal representations. Finally, the Feature Stability Enhancement (FSE) module strengthens the robustness of the disentangled but fragile representations by modeling both spatial dependencies and inter-channel correlations, thereby enabling reliable gait recognition across views and modalities.

\subsection{Multi-grained Feature Encoder}
\subsubsection{Visual Encoder}
In line with prior methodologies~\cite{guo2025camera-crossmodal-CLGait,wang2024cross-crossmodal-CrossGait,fan2023opengait-deeplearning-opengait,fan2025opengait-tpami}, we preprocess RGB and point clouds data into silhouettes and depth images, respectively. The input data is denoted as 
\begin{equation}
    X=\left \{ x^{2d}_{i},x^{3d}_i,y_{i}|i=1,2,\dots ,n \right \},
\end{equation}
where $n$ represents the number of sequences in the training set and $y_i$ denotes the identity label. Each sample $x^m_i \in \mathbb{R}^{s \times h \times w \times c}$ represents a sequence composed of $s$ frames,  $h\times w$ denotes the spatial resolution of each frame, and $c$ is the number of input channels. For global coarse-grained feature extraction, we freeze the CLIP visual encoder and add an adapter for task-specific fine-tuning:
\begin{equation}
    g^{m}_{i}= \mathrm{CLIP}_{v}\left ( x^{m}_{i} \right ).
\end{equation}
Here, $g^{m}_{i}\in\mathbb{R}^{s\times (1+o)\times d } $ consists of two parts: the [CLS] token and the remaining $o$ tokens, where $d$ is the feature dimension. Then, $g^{m}_{i}$ is passed into the adapter for task-specific fine-tuning:
\begin{equation}
    \hat{g} ^{m}_{i}= \mathrm{Adapter}\left ( \mathrm{LN} \left (  g^{m}_{i} \right )  \right ),
\end{equation}
where $\mathrm{LN}$ refers to the Layer Normalization, and the adapter consists of two $\mathrm{MLP}$ layers. Subsequently, temporal aggregation is performed using the $\mathrm{Maxpool}$ operation:
\begin{equation}
    \tilde{g}^{m}_{i} = \mathop{\mathrm{ Maxpool}} \limits_{j=1,2,\dots ,s}\left ( \hat{g} ^{m}_{i,j}   \right ),
\end{equation}
where $\tilde{g}^{m}_{i} \in \mathbb{R}^{\left ( 1+o \right ) \times d} .$
Meanwhile, to compensate for CLIP's limitations in fine-grained feature extraction, an auxiliary ResNet branch is employed to capture detailed gait information. And the same temporal aggregation is applied. The resulting local spatiotemporal representation is denoted as $\tilde{f}^{m}_{i} \in \mathbb{R}^{{h}'\times {w}'\times d}$, where ${h}'$ and ${w}'$ indicate the height and width of the feature map.

\subsubsection{Multi-grained Fusion Module}
\begin{figure}[t]
\centering
\includegraphics[width=\linewidth]{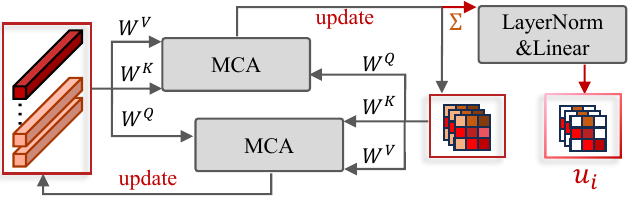}
\caption{Details of the MF module.}
\label{fig::mf}
\vspace{-0.3cm}
\end{figure}

The Multi-grained Fusion (MF) module is employed to integrate global and local representations. The features extracted from the ResNet branch are flattened into $\bar{f}^{m}_{i} \in \mathbb{R}^{  {h}'{w}' \times d} $. Subsequently, a Multi-head Cross-Attention (MCA) mechanism is applied at the $l$-th layer of the MF module to perform feature fusion as follows:
\begin{equation}
    \text{MCA}(\tilde{g}^{m}_{i(l-1)}, \bar{f}^{m}_{i(l-1)}) = \text{Concat}(\text{head}_1, \dots, \text{head}_H)W^O,
\end{equation}
\begin{equation}
    \text{head}_j = \varphi \left( \frac{( \tilde{g}^{m}_{i(l-1)} W^Q_j ) ( \bar{f}^{m}_{i(l-1)} W^K_j )^\top}{\sqrt{d_h}} \right) ( \bar{f}^{m}_{i(l-1)} W^V_j )
\end{equation}
Here, $\varphi$ represents Softmax; $W^Q_j$, $W^K_j$, and $W^V_j \in \mathbb{R}^{d \times d_h}$ denote the linear projection matrices for the $j$-th attention head; $d_h = d / H$ denotes the dimensionality of each attention head, where $H$ is the total number of heads; $W^O \in \mathbb{R}^{H d_h \times d}$ represents the linear projection matrix used to aggregate outputs from all heads. $\tilde{g}^{m}_{i(l-1)}$ and $\bar{f}^{m}_{i(l-1)}$ denote the global and local features from the $(l{-}1)$-th layer. Meanwhile, we adopt a bidirectional fusion strategy to update the global and local features at the $l$-th layer:
\begin{equation}
    \tilde{g}^{m}_{i(l)}=\text{MCA}(\tilde{g}^{m}_{i(l-1)}, \bar{f}^{m}_{i(l-1)}),
\end{equation}
\begin{equation}
    \bar{f}^{m}_{i(l)}=\text{MCA}(\bar{f}^{m}_{i(l-1)},\tilde{g}^{m}_{i(l-1)}).
\end{equation}
Ultimately, we aggregate multi-level fine-grained representations to construct the comprehensive feature embedding:
\begin{equation}
    u^{m}_{i}=\mathrm{Reshape} \left ( \mathrm{Linear} \left ( \mathrm{LN}\left (  {\textstyle \sum_{l=1}^{n_{l}}}\left ( \bar{f}^{m}_{i(l)} \right )   \right )   \right )\right ) , 
\end{equation}
where $\mathrm{Linear}$ denotes the projection layer, $n_l$ is the number of layers in the MF module, and $u^{m}_{i} \in \mathbb{R}^{{h}' \times {w}' \times d}$ is the final fused feature. The details are illustrated in Figure~\ref{fig::mf}.

\subsubsection{Spatial Weighting Module}
In addition, the Spatial Weighting (SW) module is designed to emphasize identity-discriminative regions adaptively while suppressing irrelevant areas. The spatial attention weights are computed as follows:
\begin{equation}
    w^{m}_{i}=\mathrm{C}_{1\times 1} \left ( \sigma \left (  \mathrm{BN}\left (  \mathrm{C}_{1\times 1} \left ( u^{m}_{i} \right )  \right )\right )    \right ) ,
\end{equation}
Here, $\mathrm{C}_{1\times 1}$ denotes a 2D convolution with a kernel size of $1 \times 1$, $\sigma$ refers to the LeakyReLU activation function, $\mathrm{BN}$ stands for Batch Normalization, and $w^{m}_{i} \in \mathbb{R}^{{h}' \times {w}' \times 1}$. Next, the spatial weight map $w^{m}_{i}$ is broadcasted to match the dimensionality of $u^{m}_{i}$, and used to recalibrate the feature representation:
\begin{equation}
    \tilde{u}^{m}_{i} = w^{m}_{i}\odot u^{m}_{i},
\end{equation}
where $\odot$ denotes the Hadamard product.

\subsection{Gait Modality Text Dictionary}
Initially, we design prompts to guide LLMs in generating raw textual descriptions of gait across both camera and LiDAR modalities. To mitigate the impact of viewpoint variation, we follow conventional practice by dividing viewpoints into eight distinct directions~\cite{shen2023lidargait-3d-lidargait,shen2024comprehensive-gait-survey}. Subsequently, we employ ChatGPT~\cite{achiam2023gpt} to augment the raw textual descriptions with explicit viewpoint information, thereby producing a set of modality- and view-aware semantic descriptions. To enrich diversity and ensure better compatibility with the CLIP pretraining format, we employ the multi-turn strategy to expand the original set into $l$ distinct groups of gait descriptions, which are subsequently incorporated into the Gait Modality Text Dictionary (GMTD), as illustrated in Figure~\ref{fig::prompt}. The GMTD contains $m\times 8\times l$ entries, formally defined as:
\begin{equation}
    \text{GMTD}=\left \{ t^{m}_{j} |m\in\left \{2d,3d\right \} ,j=1,2,\dots ,8l  \right \} ,
\end{equation}
where $m$ denotes different modalities.

Subsequently, for the extraction of textual features $t^{m}_{i}$ in the GMTD, we employ a frozen, pretrained CLIP text encoder followed by a two-layer MLP adapter, which yields the modality-specific textual embeddings:
\begin{equation}
    \text{GMTF}=\left \{ v^{m}_{j} |m\in\left \{2d,3d\right \} ,j=1,2,\dots ,8l  \right \} ,
\end{equation}
where $v^{m}_{j}\in \mathbb{R}^{1\times d} $. This process is described in detail in the Supplementary Material. 

\subsection{Text-guided Feature Disentanglement} 
\begin{figure}[t]
\centering
\includegraphics[width=\linewidth]{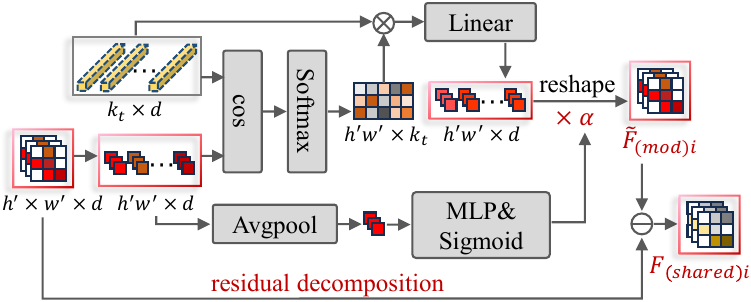}
\caption{The flowchart of the TFD module.}
\label{fig::tfd}
\vspace{-0.3cm}
\end{figure}

Upon obtaining modality-specific textual embeddings and multi-grained gait visual features, a Text-guided Feature Disentanglement (TFD) module is designed to reconstruct modality-specific gait features within the visual space, while modality-shared representations are obtained through residual decomposition. Specifically, we first select the $top\text{-}k_t$ modality-specific semantic prototypes $V_{i}^{m}=\left \{ v^{m}_{j} |j=1,2,\dots ,  k_{t} \right \}$ from the GMTF based on their cosine similarity scores with the CLIP-derived [CLS] feature ${\tilde{g*}}^{m}_{i}$. The similarity score is formally defined as:
\begin{equation}
    \text{cos}({\tilde{g*}} ^{m}_{i}, v^{m}_{j}) = \frac{{\tilde{g*}} ^{m}_{i} \cdot v^{m}_{j} }{\|{\tilde{g*}} ^{m}_{i}\|_2 \cdot \|v^{m}_{j}\|_2}.
\end{equation}
Subsequently, the modality text group $V_{i}^{m}$ is employed to reconstruct modality-specific information. In particular, each prototype in $V_{i}^{m}$ is first projected into a shared latent space, followed by L2 normalization, yielding $\widehat{V}_{i}^{m}\in\mathbb{R}^{k_{t}\times d} $. Meanwhile, $\tilde{u}^{m}_{i}$ is flattened into $\bar{u}^{m}_{i} \in \mathbb{R}^{{h}'{w}' \times d}$, and normalized along the feature dimension to obtain $\hat{u}^{m}_{i}$. Next, we compute the affinity weights based on cosine similarity, followed by a Softmax operation to normalize the map:
\begin{equation}
    \Omega  = \varphi  \left( \mathrm{cos} \left ( \hat{u}^{m}_{i},\widehat{V}_{i}^{m} \right )  \right),
\end{equation}
where $\Omega \in \mathbb{R}^{{h}'{w}' \times k_{t}} $. We further perform a weighted fusion of the modality-specific semantic features according to the attention map, thereby reconstructing modality-specific features in the visual space:
\begin{equation}
    {F_{(mod)}}^{m}_{i} = \text{Reshape}\left( \mathrm{Linear} \left ( \Omega \widehat{V}_{i}^{m} \right )    \right) ,
\end{equation}
where ${F_{(mod)}}^{m}_{i} \in \mathbb{R}^{{h}'\times {w}'\times d}.$ To control the influence of ${F_{(mod)}}^{m}_{i}$ during the early stages of training and prevent model divergence, we introduce a gating mechanism that adaptively modulates its contribution. Formally,
\begin{equation}
    \alpha = \eta \left (   \mathrm{MLP}\left ( \mathop{\mathrm{ Avgpool}} \limits_{{h}'\times {w}'  }\left ( \tilde{u}^{m}_{i}  \right ) \right ) \right ),
\end{equation}
\begin{equation}
    {\widetilde{F}_{(mod)}}{}^{m}_{i}=\alpha \odot {F_{(mod)}}^{m}_{i}.
\end{equation}
Here, $\mathrm{MLP}$ denotes a multilayer perceptron, $\eta$ represents the Sigmoid function, $\alpha \in\mathbb{R} ^{1\times d}$ is the channel-wise modulation factor. Furthermore, to obtain modality-shared representations, we adopt a residual decomposition strategy as follows:
\begin{equation}
    {F}_{(shared)}{}^{m}_{i} = \tilde{u}^{m}_{i}-{\widetilde{F}_{(mod)}}{}^{m}_{i},
\end{equation}
where ${F}_{(shared)}{}^{m}_{i}\in \mathbb{R}^{{h}'\times {w}'\times d} $ denotes the modality-shared gait representation. The flowchart is shown in Figure~\ref{fig::tfd}

\subsection{Feature Stability Enhancement}
\begin{figure}[t]
\centering
\includegraphics[width=\linewidth]{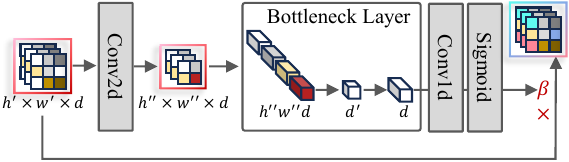}
\caption{Illustration of the FSE module.}
\label{fig::fse}
\vspace{-0.3cm}
\end{figure}

Although the modality-shared features have been semantically disentangled from modality-specific components via the TFD module, the remaining modality-specific noise and local perturbations in the original features still limit their stability and discriminative power. To mitigate this issue, we introduce a Feature Stability Enhancement (FSE) module, which is designed to capture local spatial receptive fields and global channel-wise dependencies, thereby enhancing the robustness and discriminability of modality-shared representations, as illustrated in Figure~\ref{fig::fse}. Specifically, we first model the local spatial correlations:
\begin{equation}
    \hat{F}_{(shared)}{}^{m}_{i}  = \mathrm{C}_{3\times 3} \left ( {F}_{(shared)}{}^{m}_{i} \right ) ,
\end{equation}
where $\hat{F}_{(shared)}{}^{m}_{i} \in \mathbb{R} ^{{h}''\times {w}''\times d }.$

Subsequently, a bottleneck layer is utilized to transform the channel-wise features, wherein the downsampled feature map is first flattened and projected into a compact ${d}'$-dimensional latent space(${d}'\ll d$), followed by an inverse projection to recover the original channel dimensionality:
\begin{equation}
    \bar{F}_{(shared)}{}^{m}_{i} = W_{2} \left (  \sigma \left (   W_{1}\left ( \mathrm{Flatten}\left ( \hat{F}_{(shared)}{}^{m}_{i} \right )  \right ) \right )\right ) ,
\end{equation}
where $W_{1} \in \mathbb{R}^{{h}''{w}'' d\times {d}' } ,W_{2} \in \mathbb{R}^{{d}' \times d },$ and $\bar{F}_{(shared)}{}^{m}_{i} \in \mathbb{R}^{1\times d} $. In modality-shared features, different channel dimensions often encode distinct cross-modal semantic patterns; thus, modeling the inter-channel dependencies and performing adaptive weighting is essential for improving feature robustness. Specifically, a 1D convolution is employed to capture dependencies, followed by a Softmax operation to compute the weight of each channel dimension:
\begin{equation}
    \beta =\eta \left ( \mathrm{C}_{3}\left ( \bar{F}_{(shared)}{}^{m}_{i}  \right )  \right ) ,
\end{equation}
where $\beta \in \mathbb{R} ^{1\times d}$ and $\mathrm{C}_{3}$ denotes a 1D convolution with a kernel size of 3. Subsequently, we adjust the channel weights to obtain features based on global dependency correlations:
\begin{equation}
    \widetilde{F}_{(shared)}{}^{m}_{i} = \beta \odot {F}_{(shared)}{}^{m}_{i}.
\end{equation}
Finally, a combination of HPP\cite{chao2021gaitset-tpami-hpp} and Separated FCs\cite{fan2023opengait-deeplearning-opengait} is utilized to extract part-based discriminative features, denoted as ${F_{(*)}}^{m}_{i} \in \mathbb{R}^{n_{p}\times d} $, where $n_p$ denotes the number of parts.

\subsection{Loss Functions}
To explicitly enforce structural disentanglement between modality-specific and modality-shared features, and to further strengthen cross-modal representation learning guided by textual semantics, we propose three collaboratively optimized loss functions: Modality Alignment Loss (MA Loss), Modality Orthogonality Loss (MO Loss), and HSIC-based Independence Loss (HSIC Loss). Collectively, these losses impose complementary constraints on the network from the perspectives of semantic alignment, feature disentanglement, and statistical decorrelation, establishing a unified framework of alignment, orthogonality, and independence.
\textbf{MA Loss} is introduced to ensure that the reconstructed feature accurately reflects the modality-specific semantic priors encoded in the GMTF. The specific formulation is as follows:
\begin{equation}
    \mathcal{L}^{m}_{align}= 1 - \frac{1}{N}  \sum_{i=1}^{N} \left ( \mathrm{cos}\left (  {\bar{F}_{(mod)}}{}^{m}_{i}, \bar{V}{}_{i}^{m} \right )  \right ) .
\end{equation}
Here, $N$ denotes the number of batch size, ${\bar{F}_{(mod)}}{}^{m}_{i} \in \mathbb{R}^{1\times d} $ refers to the global average pooled vector of the modality-specific ${\widetilde{F}_{(mod)}}{}^{m}_{i}$, and $\bar{V}{}_{i}^{m} \in \mathbb{R}^{1\times d} $ denotes the mean of the corresponding text embeddings $\widehat{V}{}_{i}^{m}$ for each sample.

\textbf{MO Loss} is designed to achieve effective disentanglement between the modality-specific feature ${\widetilde{F}_{(mod)}}{}^{m}_{i}$ and the modality-shared feature $\widetilde{F}_{(shared)}{}^{m}_{i}$, which encourages the two representations to be mutually independent in the embedding space and is defined as follows:
\begin{equation}
    \mathcal{L}^{m}_{ortho}= \frac{1}{N}  \sum_{i=1}^{N}  \left ( \frac{\left \langle \widetilde{F}_{(shared)}{}^{m}_{i}, {\widetilde{F}_{(mod)}}{}^{m}_{i} \right \rangle }{\| \widetilde{F}_{(shared)}{}^{m}_{i} \|  \| {\widetilde{F}_{(mod)}}{}^{m}_{i} \|}\right ).
\end{equation}

\begin{table*}[t] \footnotesize
\caption{Rank-$1$ accuracy of cross-modal gait recognition from 2D camera to 3D LiDAR on the SUSTech1K dataset. $^{\spadesuit}$ indicates that a significant quantity of extra data is required for pre-training in this approach. The best results are indicated in \textbf{bold}, second in \underline{underline}.} 
\centering
\resizebox{\textwidth}{!}{
\begin{tabular}{ll|cccccccc|cc}
\toprule
\multicolumn{2}{c|}{\multirow{2}{*}{\textbf{Methods}}} & \multicolumn{8}{c}{\textbf{Camera(2D)} $\to$ \textbf{LiDAR(3D)}} & \multicolumn{2}{c}{\textbf{Overall}} \\
\cmidrule(r){3-12}
& & Normal & Bag & Clothing & Carrying & Umbrella & Uniform & Occlusion & Night & \textbf{Rank-$1$} & \textbf{Rank-$5$} \\
\midrule
CAJ~\cite{ye2021channel-exp-CAJ} & ICCV'21 & 16.4 & - & 7.5 & - & 7.4 & - & - & 2.4 & 11.3 & 30.1 \\
SAAI~\cite{fang2023visible-exp-SAAI} & ICCV'23 & 22.4 & - & 14.3 & - & 14.0 & - & - & 5.3 & 23.1 & 49.5 \\
LidarGait~\cite{shen2023lidargait-3d-lidargait} & CVPR'23 & 18.2 & - & 3.4 & - & 3.4 & - & - & 4.7 & 9.6 & 28.1 \\
CL-Gait$^{\spadesuit}$~\cite{guo2025camera-crossmodal-CLGait} & ECCV'24 & - & - & - & - & - & - & - & - & \underline{55.1} & 77.3 \\
CrossGait~\cite{wang2024cross-crossmodal-CrossGait} & IJCB'24 & \underline{63.2} & - & \underline{30.6} & - & 38.5 & - & - & \textbf{11.8} & 53.6 & 77.0 \\
IDKL~\cite{ren2024implicit-IDKL} & CVPR'24 & 60.3 & 49.8 & 29.2 & 48.5 & 36.9 & 50.7 & 64.2 & 9.4 & 52.2 & 75.2 \\
TVI-LFM~\cite{hu2024empowering-TVILFM} & NeurIPS'24 & 61.0 & 50.3 & 30.1 & 50.2 & 37.5 & 51.0 & 66.5 & 10.0 & 53.0 & 76.1 \\

TSKD~\cite{shi2026two-Exp-PR2025-VIReID-TSKD} & PR'25 & 52.1 & 43.6 & 27.9 & 48.0 & 32.7 & 41.1 & 55.6 & 6.3 & 42.6 & 65.8 \\

SCR~\cite{yu2025no-Exp-IF2025-VIReID-SCR} & IF'25 & 61.3 & \underline{52.9} & 29.6 & \underline{53.0} & 39.1 & \textbf{53.7} & \underline{69.4} & 10.3 & 54.9 &78.1 \\

\rowcolor{gray!20}
\textbf{TCFDNet} & \textbf{Ours} 
& \textbf{67.6}
& \textbf{60.8}
& \textbf{36.1}
& \textbf{55.4}
& 39.1
& \underline{52.8}
& \textbf{71.2}
& \underline{11.2}
& \textbf{55.9}
& 78.1 \\
\bottomrule
\end{tabular}
}
\label{tab::sustech1k2d3d}
\vspace{-0.2cm}
\end{table*}

\begin{table*}[t]
\caption{Rank-$1$ accuracy of cross-modal gait recognition from 3D LiDAR to 2D camera on the SUSTech1K dataset. $^{\spadesuit}$ denotes methods pre-trained on synthetic data.}
\centering
\resizebox{\textwidth}{!}{
\begin{tabular}{ll|cccccccc|cc}
\toprule
\multicolumn{2}{c|}{\multirow{2}{*}{\textbf{Methods}}} & \multicolumn{8}{c}{\textbf{LiDAR(3D)} $\to$ \textbf{Camera(2D)}} & \multicolumn{2}{|c}{\textbf{Overall}} \\
\cmidrule(r){3-12}
& & Normal & Bag & Clothing & Carrying & Umbrella & Uniform & Occlusion & Night & \textbf{Rank-$1$} & \textbf{Rank-$5$} \\
\midrule
CAJ~\cite{ye2021channel-exp-CAJ} & ICCV'21 & 15.3 & - & 6.4 & - & 13.0 & - & - & 2.3 & 12.3 & 32.3 \\
SAAI~\cite{fang2023visible-exp-SAAI} & ICCV'23 & 26.5 & - & 21.9 & - & 23.2 & - & - & 3.2 & 26.1 & 54.1 \\
LidarGait~\cite{shen2023lidargait-3d-lidargait} & CVPR'23 & 23.2 & - & 14.2 & - & 24.7 & - & - & 2.4 & 18.3 & 39.6 \\
CL-Gait$^{\spadesuit}$~\cite{guo2025camera-crossmodal-CLGait} & ECCV'24 & - & - & - & - & - & - & - & - & 53.3 & 75.6 \\
CrossGait~\cite{wang2024cross-crossmodal-CrossGait} & IJCB'24 & \underline{62.2} & - & 35.4 & - & 57.8 & - & - & \underline{10.3} & 56.4 & \underline{79.8} \\
IDKL~\cite{ren2024implicit-IDKL} & CVPR'24 & 59.6 & 52.3 & 31.0 & 49.5 & 55.2 & \underline{56.1} & 65.3 & 7.9 & 54.8 & 77.1 \\
TVI-LFM~\cite{hu2024empowering-TVILFM} & NeurIPS'24 & 60.4 & 53.0 & 32.7 & 51.6 & 56.4 & 55.8 & 69.2 & 9.1 & 55.7 & 78.5 \\

TSKD~\cite{shi2026two-Exp-PR2025-VIReID-TSKD} & PR'25 & 50.1 & 41.3 & 27.7 & 42.8 & 45.9 & 46.2 & 52.5 & 7.8 & 47.2 & 68.1 \\

SCR~\cite{yu2025no-Exp-IF2025-VIReID-SCR} & IF'25 & 61.6 & \underline{54.1} & \underline{35.8} & \underline{52.0} & \underline{58.1} & 55.9 & \underline{72.6} & 10.2 & \underline{57.7} & 79.5 \\

\rowcolor{gray!20}
\textbf{TCFDNet} & \textbf{Ours} 
& \textbf{70.9}
& \textbf{64.8}
& \textbf{36.7}
& \textbf{59.1}
& \textbf{63.3}
& \textbf{64.3}
& \textbf{78.7}
& \textbf{11.1}
& \textbf{61.7}
& \textbf{82.5} \\
\bottomrule
\end{tabular}}
\label{tab::sustech1k3d2d}
\vspace{-0.2cm}
\end{table*}

\textbf{HSIC Loss} is employed to decorrelate the distributions of the two feature representations from a statistical dependence perspective:
\begin{equation}
    \mathcal{L}^{m}_{\text{HSIC} }= \mathrm{HSIC}\left ( {\widetilde{F}_{(mod)}}{}^{m}_{i}, \widetilde{F}_{(shared)}{}^{m}_{i}\right ) = \frac{\mathrm{tr}\left ( K_{c}L_{c} \right )   }{\left ( N-1 \right )^{2} } .
\end{equation}
Here, $K_{c},L_{c} \in \mathbb{R}^{N\times N} $ denote the centered linear kernel Gram matrices of ${\widetilde{F}_{(mod)}}{}^{m}_{i}$ and $\widetilde{F}_{(shared)}{}^{m}_{i}$, respectively.

In addition, we adopt the conventional triplet loss $\mathcal{L}_{tri} $ and cross-entropy loss $\mathcal{L}_{ce}$ to optimize the model~\cite{fan2023opengait-deeplearning-opengait,wang2024cross-crossmodal-CrossGait}. Detailed analysis of these loss functions is provided in the Supplementary Material. In summary, the overall loss function is defined as:

\begin{equation}
\begin{aligned}
    \mathcal{L}_{all} =\gamma_{1} \left ( \mathcal{L}_{tri}+ \mathcal{L}_{ce}\right ) &+ \gamma_{2}\left ( \mathcal{L}^{m}_{align} \right ) \\ &+ \gamma_{3}\left ( \mathcal{L}^{m}_{ortho} + \mathcal{L}^{m}_{\text{{HSIC}} } \right )  .
\end{aligned}
\end{equation}
By default, $\gamma_{1}=1.0$, $\gamma_{2}=0.5$, and $\gamma_{3}=0.1$.
\section{Experiments}
\subsection{Dataset}

\begin{table}[t] \footnotesize
\caption{Accuracy of cross-modal gait recognition on the FreeGait.} 
\centering
\resizebox{\linewidth}{!}{
\begin{tabular}{ll|cccc}
\toprule
\multicolumn{2}{c|}{\multirow{2}{*}{\textbf{Methods}}} & \multicolumn{2}{c}{\textbf{2D $\to$ 3D}} & \multicolumn{2}{c}{\textbf{3D $\to$ 2D}} \\
\cmidrule{3-6}
 & & \textbf{R-$1$} & \textbf{R-$5$} & \textbf{R-$1$} & \textbf{R-$5$} \\
\midrule
HMRNet\cite{han2024gait-3d-freegait} & MM'24 & 23.5 & 55.7 & 25.1 & 57.0 \\
CrossGait\cite{wang2024cross-crossmodal-CrossGait} & IJCB'24 & 29.6 & 60.8 & 32.3 & 65.9 \\
IDKL~\cite{ren2024implicit-IDKL} & CVPR'24 & 36.7 & 67.4 & 39.5 & 70.3 \\
TVI-LFM~\cite{hu2024empowering-TVILFM} & NeurIPS24 & 38.9 & 69.1 & 41.0 & 71.8 \\

TSKD~\cite{shi2026two-Exp-PR2025-VIReID-TSKD} & PR'25 & 25.1 & 57.9 & 26.7 & 60.8 \\

SCR~\cite{yu2025no-Exp-IF2025-VIReID-SCR} & IF'25 & \underline{40.1} & \underline{72.0} & \underline{43.3} & \underline{75.9} \\

\rowcolor{gray!20}
\textbf{TCFDNet} & \textbf{Ours}
& \textbf{52.1} & \textbf{85.3} 
& \textbf{57.9} & \textbf{87.2} \\

\bottomrule
\end{tabular}
}

\label{tab::comparsionFreeGait}
\vspace{-0.3cm}
\end{table}

\textbf{SUSTech1K}~\cite{shen2023lidargait-3d-lidargait} comprises 25,239 gait sequences from 1,050 subjects, captured using synchronized RGB cameras and LiDAR sensors across 12 viewpoints and 8 walking conditions (e.g., clothing changes, carrying). Following standard protocol, we use 6,011 sequences from 250 identities for training and 19,228 sequences from 800 unseen identities for testing. For cross-modal evaluation, sequences under varied conditions in one modality serve as probes, while their normal-condition counterparts in the other modality form the gallery. Recognition performance is further reported across multiple covariate subsets for detailed analysis. 

\textbf{FreeGait}~\cite{han2024gait-3d-freegait} includes 11,921 sequences from 1,195 individuals, collected under unconstrained outdoor environments using high-resolution 128-line OUSTR LiDAR and RGB cameras from three viewpoints within a 25-meter radius. The dataset features diverse conditions, including 51 subjects recorded under low-light scenarios. We follow the original evaluation split, using 5,000 sequences from 500 identities for training and 6,921 sequences from the remaining 695 identities for testing.


\subsection{Implementation Details}
To ensure fair comparisons with prior methods~\cite{guo2025camera-crossmodal-CLGait,wang2024cross-crossmodal-CrossGait,shen2025lidargait++-cvpr}, we follow their input settings by using camera silhouettes and LiDAR depth maps, both resized to $64 \times 64$. 
Regarding the network architecture, we employ a trainable ResNet-9 as the MFE adapter, with the embedding dimension $d$ set to 512 and the number of part tokens $o$ fixed at 16.
During training, we randomly sample 10 frames per sequence, with a batch size of 8 (identities) $\times$ 8 (sequences). The proposed TCFDNet is optimized using the Adam optimizer with an initial learning rate of $3 \times 10^{-4}$, and trained for a total of 25{,}000 epochs. A MultiStepLR scheduler is employed to decay the learning rate by a factor of 0.1 at the 10{,}000th and 20{,}000th epochs. Our implementation is based on the OpenGait~\cite{fan2023opengait-deeplearning-opengait,fan2025opengait-tpami} framework, and we adopt the Patch Exchange data augmentation strategy, which is detailed in the supplementary material. All experiments are conducted on two NVIDIA RTX 3090 GPUs.












\begin{table}[t]
\centering
\caption{Ablation study on SUSTech1K dataset for cross-modal gait recognition (LiDAR $\rightarrow$ Camera). At each step, only one functional group is modified while others remain fully integrated.}
\resizebox{\linewidth}{!}{
\begin{tabular}{c|cccc|cc|cc}
\toprule

\textbf{Text} & \multicolumn{4}{c|}{\textbf{Visual Backbone}} & \multicolumn{2}{c|}{\textbf{Decoupling}} & \multicolumn{2}{c}{\textbf{Overall}}  \\
\midrule

GMTD & ViT & ResNet & MF & SW & TFD & FSE & \textbf{R-$1$} & \textbf{R-$5$}\\
\midrule

$\times$ & \multicolumn{4}{c|}{\textit{full integration}} & \multicolumn{2}{c|}{\textit{full integration}} & 56.2 & 77.3 \\
\midrule

\checkmark & \checkmark & $\times$ & $\times$ & $\times$ & \multicolumn{2}{c|}{\multirow{3}{*}{\textit{full integration}}} & 54.9 & 74.6 \\

\checkmark & $\times$ & \checkmark & $\times$ & \checkmark & & & 58.4 & 78.9 \\

\checkmark & $\times$ & \checkmark & $\times$ & $\times$ & & & 56.7 & 76.5 \\

\midrule

\checkmark & \multicolumn{4}{c|}{\multirow{2}{*}{\textit{full integration}}} & \checkmark & $\times$ & 59.8 & 80.6 \\

\checkmark & & & & & $\times$ & $\times$ & 58.9 & 79.3 \\
\midrule
\rowcolor{gray!20}
\checkmark & \checkmark & \checkmark & \checkmark & \checkmark & \checkmark & \checkmark & \textbf{61.7} & \textbf{82.5} \\

\bottomrule
\end{tabular}
}
\label{tab:grouped_ablation_lidar2camera}
\vspace{-0.3cm}
\end{table}

\subsection{Comparative Results}
\begin{figure}[t]
\centering
\includegraphics[width=\linewidth]{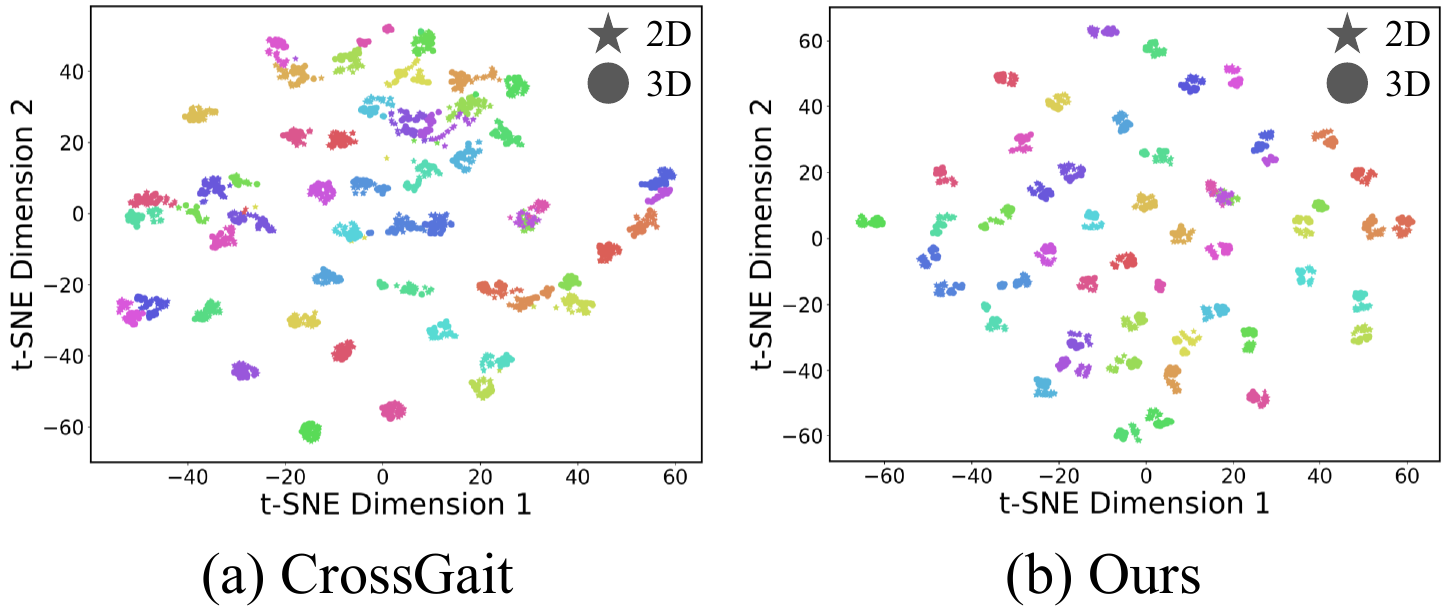}
\caption{t-SNE visualization of cross-modal 2D and 3D features. Zooming in for details.}
\label{fig::tsne}
\vspace{-0.3cm}
\end{figure}

\begin{figure}[t]
\centering
\includegraphics[width=\linewidth]{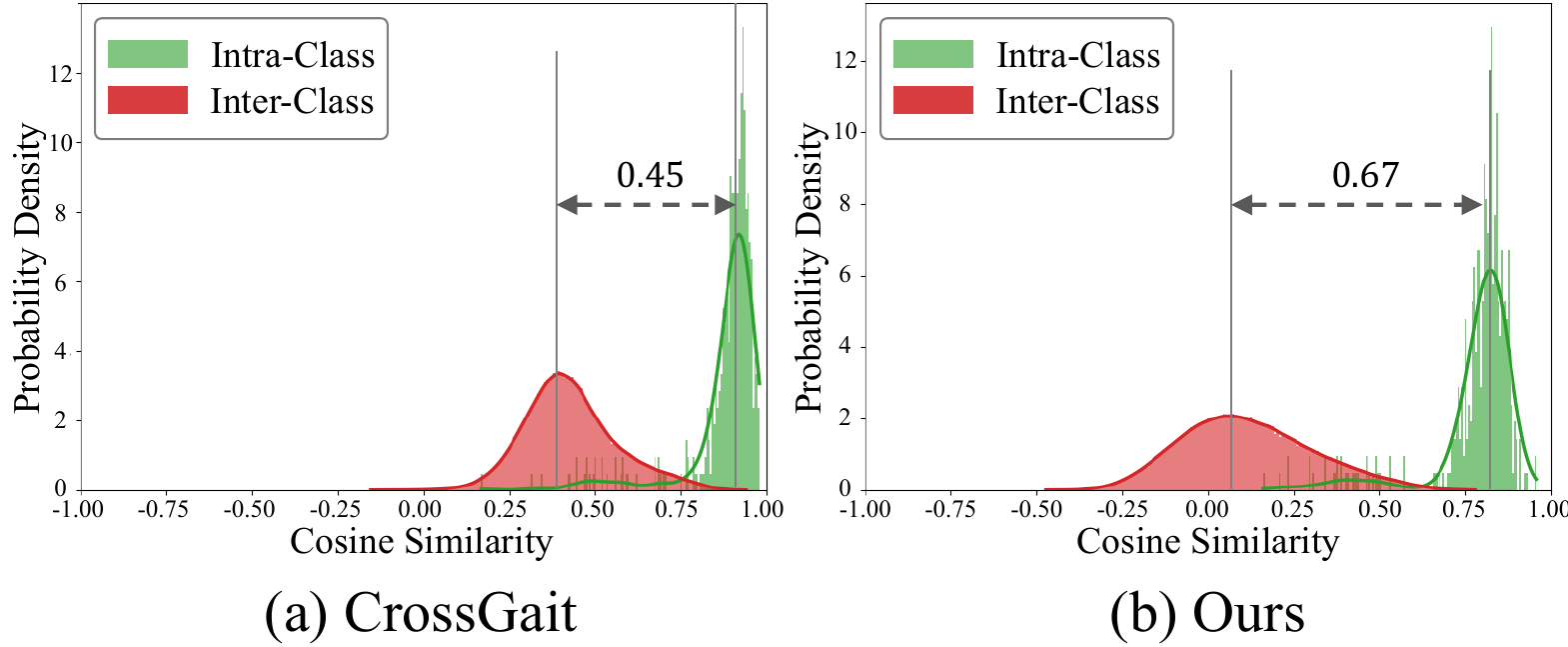}
\caption{Visualization of cross-modal intra/inter-class cosine similarity distribution.}
\label{fig::distribution}
\vspace{-0.3cm}
\end{figure}

We conduct comprehensive comparisons against state-of-the-art (SOTA) methods on the LCCGait benchmark and further adapt leading approaches from the Visible-Infrared Re-Identification task to establish a more rigorous experimental baseline. On the SUSTech1K dataset, the proposed TCFDNet consistently outperforms previous state-of-the-art (SOTA) methods under both 2D$\to$3D and 3D$\to$2D retrieval settings. The results are presented in Table~\ref{tab::sustech1k2d3d} and Table~\ref{tab::sustech1k3d2d}. Notably, our approach maintains robust and effective cross-modal retrieval performance even in challenging scenarios such as Clothing and Occlusion. However, the performance of TCFDNet degrades under Night conditions, which is a common limitation among most existing methods. We attribute this to the newly introduced day–night cross-domain and cross-modal discrepancy, which represents a promising direction for future research. In addition, we visualize the results of different comparison methods using t-SNE~\cite{van2008visualizing-tsne} and cross-modal similarity distributions~\cite{shi2024multi-vireid-distribution}, as shown in Figure~\ref{fig::tsne} and Figure~\ref{fig::distribution}. These visualization results further highlight that TCFDNet effectively captures modality-invariant semantics while preserving strong discriminative capability, leading to more compact intra-class distances and larger inter-class separations. On the FreeGait dataset, we reproduced prior methods and conducted detailed comparative experiments, as shown in Table~\ref{tab::comparsionFreeGait}. The results show that our method achieves SOTA performance, validating its strong generalization in real-world in-the-wild scenarios.
\begin{figure}[!htbp]
\centering
\includegraphics[width=\linewidth]{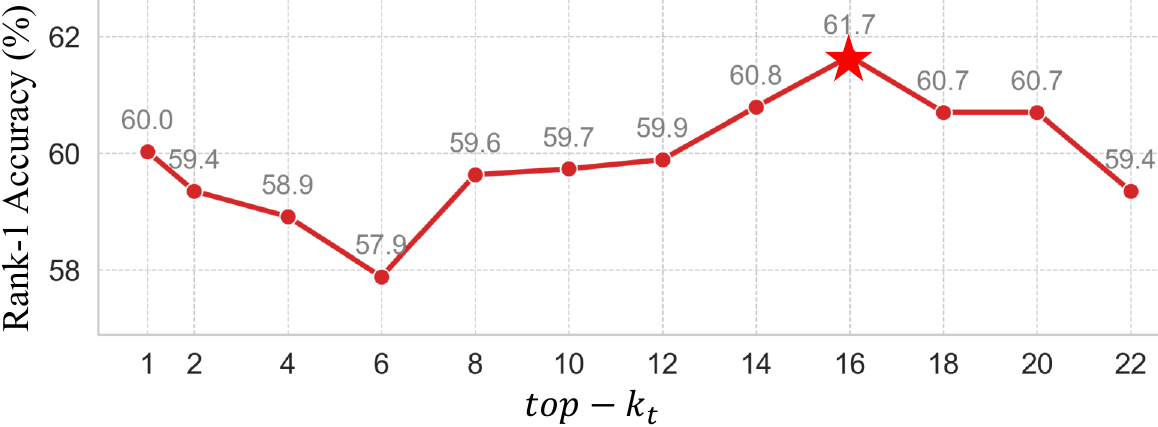}
\caption{Illustration of Rank-$1$ accuracy on the number of $top\text{-}k_t$ within the GMTD module. }
\label{fig::topkt}
\vspace{-0.5cm}
\end{figure}

\subsection{Ablation Study}
We conducted detailed ablation studies to verify the effectiveness of each proposed module, as shown in Table~\ref{tab:grouped_ablation_lidar2camera}. 
Notably, removing the GMTD module leads to a substantial performance degradation, underscoring the critical role of modality-specific textual priors in facilitating effective cross-modal feature disentanglement. Furthermore, we investigate the impact of the $top\text{-}k_t$ text prototype selection in GMTD. As reported in Figure~\ref{fig::topkt}, the best performance is observed when $k_t = 16$, which balances the semantic diversity and alignment specificity. The detailed ablation studies on the patch exchange data augmentation and the loss function design are provided in the Supplementary.
\section{Conclusion}
We introduce TCFDNet, a Text-guided Cross-modal Feature Disentanglement Network for cross-modal gait recognition. By leveraging LLMs, TCFDNet constructs modality-aware textual embeddings and aligns them with visual representations through a CLIP-based multi-grained encoder. A text-guided disentanglement module further isolates modality-shared and specific semantics, while a feature stability enhancement module improves robustness via spatial–channel dependency modeling. Three disentanglement losses jointly ensure fine-grained supervision and effective feature separation. The patch exchange augmentation is further designed to bridge the input gaps during training. Comprehensive experiments demonstrate that TCFDNet achieves superior accuracy and generalization, establishing a new paradigm for text-guided cross-modal gait recognition.
{
    \small
    \bibliographystyle{ieeenat_fullname}
    \bibliography{ref}
}


\end{document}